\documentclass[runningheads]{llncs}

\usepackage{eccv}

\usepackage{eccvabbrv}

\usepackage{graphicx}
\usepackage{amssymb}
\usepackage{booktabs}

\usepackage[accsupp]{axessibility}  

\usepackage{hyperref}

\usepackage{orcidlink}

\usepackage{url}            
\usepackage{amsfonts}       
\usepackage{nicefrac}       
\usepackage{microtype}      
\usepackage{xcolor}         
\usepackage{xspace}
\usepackage{amsmath} 
\usepackage[ruled,vlined]{algorithm2e}
\usepackage{wrapfig}
\usepackage{multirow}
\usepackage{tabularx}

\newcommand{\ourmethod}{\textbf{THE\_AMAZING\_METHOD}\xspace}
\definecolor{tanzila}{RGB}{220,20,60}
\usepackage{xcolor}

\newcommand{\up}[1]{\textcolor{blue}{(+#1)}}
\newcommand{\down}[1]{\textcolor{red}{(-#1)}}

\begin{document}

\title{Activation Outliers Matter: Robust Recovery for Quantized Multimodal LLMs} 

\titlerunning{Robust Recovery for Quantized Multimodal LLMs}


\author{Tanzila Rahman \and
Mehran Taghian Jazi \and
Yunke Peng \and
Zhuang Ma \and
Anandharaju Durai Raju \and
Yao Wang \and
Xing Huang \and
Hei Yi Mak \and
Shadan Golestan \and
Hoang Le \and
Yonghan Dong \and
Wei Guo \and
Yaoyuan Wang}

\authorrunning{T. Rahman et al.}

\institute{Huawei\\
\email{\{tanzila.rahman, pengyunke\}@huawei.com}}

\maketitle

\begin{abstract}
Low-bit quantization offers a promising avenue for reducing the computational and memory demands of Multimodal Large Language Models (MLLMs). Recent hardware support for low-precision formats, ranging from MXFP8 to ultra-low-bit formats such as MXFP4 and HiF4, has accelerated research into efficient MLLM training and deployment. In this work, we present a systematic study of these quantization schemes in representative MLLMs that span both video generation and reasoning tasks. Our analysis shows that MXFP8 achieves near-lossless performance, whereas aggressive 4-bit quantization leads to significant degradation. Through extensive ablations, we identify activation quantization as the primary source of this performance loss, contributing substantially more than weight quantization. 
Motivated by this observation, we propose \textbf{R}esidual \textbf{F}allback \textbf{Q}uantization (\textbf{RFQ}), a lightweight activation reconstruction framework that supplements the primary ulta-low-bit activation representation with an auxiliary quantized residual pathway. By explicitly modeling and compensating for quantization errors, RFQ improves activation fidelity while preserving the efficiency advantages of ultra-low-bit computation. RFQ requires no architectural modifications and incurs negligible computational overhead. Extensive experiments on Wan2.2 and Qwen3-VL demonstrate that RFQ consistently recovers a substantial portion of the performance lost under the quantization of MXFP4 and HiF4, significantly narrowing the gap to BF16 baselines across both generation and 4 reasoning benchmarks. Our findings establish activation quantization as the dominant bottleneck in ultra-low-bit MLLMs and highlight residual-based activation reconstruction as an effective and practical strategy for robust 4-bit deployment.

\end{abstract}

\section{Introduction}
Foundation Models (FMs), particularly Multimodal Large Language Models (MLLMs), have emerged as a dominant paradigm for building generalist AI systems capable of jointly reasoning over language, vision, and other modalities. This unified modeling paradigm has driven substantial progress in visual understanding~\cite{NEURIPS2024_a0303731, NEURIPS2024_cdcc6d47}, multimodal reasoning~\cite{shao2024visual, NEURIPS2024_32923dff}, and agentic interaction~\cite{agashe2025agent, szot2024grounding}. Despite these advances, the training and deployment of MLLMs remain prohibitively expensive due to their large parameter scales, long-context computation, and heterogeneous multimodal architectures, all of which impose significant memory and compute demands~\cite{xi2025coat,bhatnagar2025luq}.
To improve efficiency, a wide range of techniques have been explored, including sparse attention mechanisms~\cite{NEURIPS2020_c8512d14}, efficient decoding strategies~\cite{hooper2025speed}, and most prominently low-bit quantization~\cite{xi2025coat, frantar2023optq, MLSYS2024_42a452cb}. Quantization reduces both memory footprint and computational cost by representing model tensors, including weights, activations, gradients, and optimizer states, with reduced-precision formats such as FP8.

\begin{figure}[t]
    \centering
    \includegraphics[width=\textwidth]{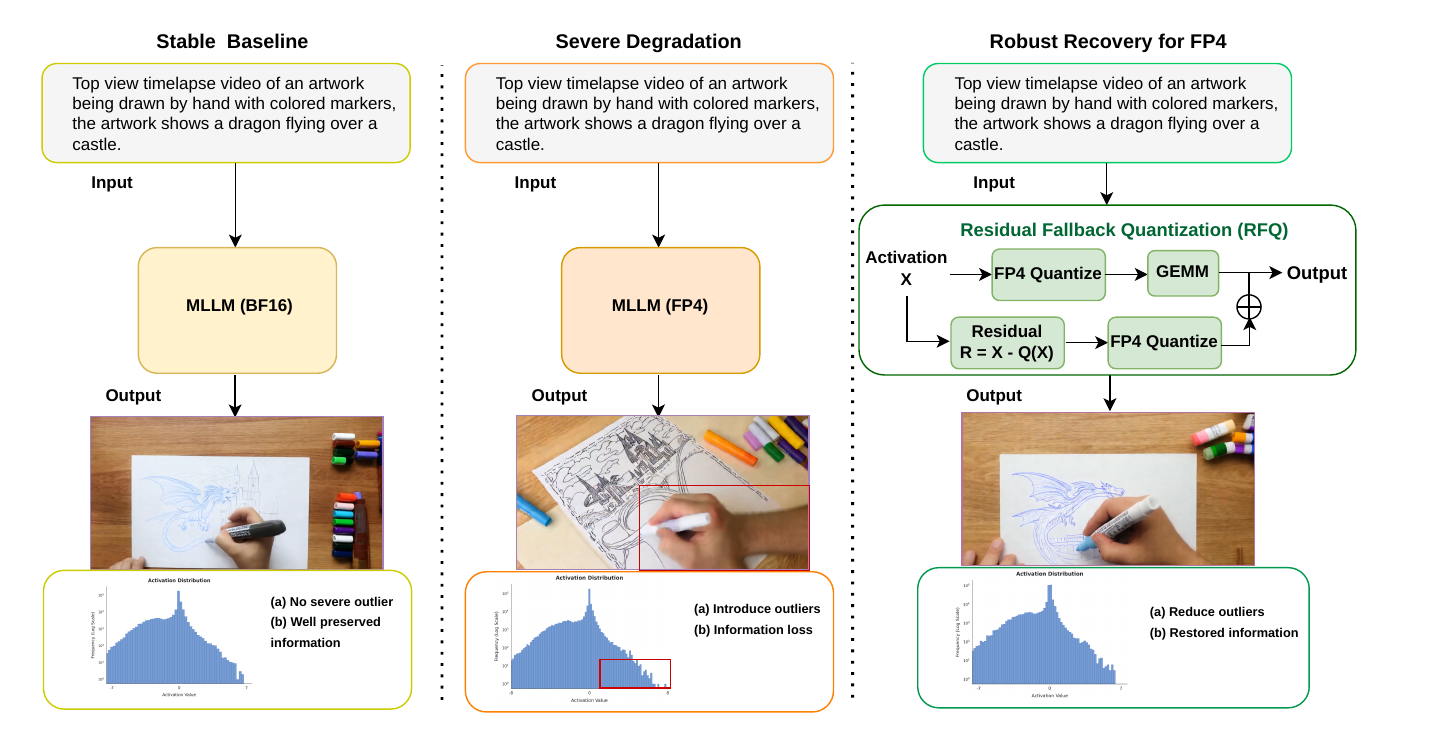}
    \vspace{-0.25in}
    \caption{Overview of the impact of activation outliers and the proposed RFQ framework. Activation outliers degrade MLLM performance under FP4 quantization. RFQ reduces outlier-induced quantization errors through residual fallback correction, recovering generation quality close to the BF16 baseline.}
    \vspace{-0.25in}
    \label{fig:teaser}
\end{figure}

Recent advances in hardware-supported mixed-precision training have significantly expanded the applicability of low-bit computation in large-scale systems. Early efforts such as TransformerEngine~\cite{nvidia2024transformerengine} demonstrated the effectiveness of matrix multiplications to accelerate linear layers. FP8-LM~\cite{peng2023fp8lmtrainingfp8large} further extended FP8 quantization to gradients, optimizer states, and communication, reducing memory and bandwidth overhead during training. More recently, COAT~\cite{xi2025coat} advanced end-to-end FP8 training by additionally quantizing activations and second-order optimizer states, while introducing tensor-specific strategies such as dynamic range expansion for optimizer statistics and fine-grained activation quantization for sensitive nonlinear layers. Alternatively, ~\cite{bondarenko2023quantizable} identifies attention-induced activation outliers in transformers and introduces clipped softmax and gated attention to suppress them, enabling full INT8 quantization without extra fine-tuning. These developments highlight that effective low-precision training requires careful and component-aware design throughout the training stack.

Motivated by these gains, we investigate whether further efficiency improvements can be achieved by pushing multimodal training into the more aggressive FP4 regime. However, the limited dynamic range and precision of FP4 introduce substantial numerical instability, often resulting in degraded convergence and accuracy. To understand this behavior, we conduct a systematic study on two representative MLLMs: Wan2.2~\cite{wan2025wanopenadvancedlargescale}, a multimodal generative model, and Qwen3-VL~\cite{bai2025qwen25vltechnicalreport}, a multimodal reasoning model. Following ~\cite{hu2026masquant}, we also quantize key components spanning vision encoders, text encoders, attention blocks, feed-forward networks, multimodal projection layers, and mixture-of-expert modules.
Our analysis reveals that FP4 degradation is highly heterogeneous across both models and modules. In Wan2.2, the most severe performance drop occurs in the visual generation backbone (\emph{i.e.} WanDiT), while the UMT5~\cite{chungunimax} text encoder remains relatively robust. In contrast, Qwen3-VL exhibits a heightened sensitivity within its vision encoder relative to the language modules; nonetheless, the language backbone also undergoes non-trivial architectural degradation under ultra-low-bit constraints. Across both models, we find that activation quantization is the primary driver of accuracy degradation, with certain modules exhibiting heavy-tailed activation distributions and pronounced outliers that are severely distorted under FP4 precision, leading to amplified downstream errors during training.

Prior work on outlier-aware quantization provides partial solutions to the activation instability. Post-training quantization methods such as SmoothQuant~\cite{xiao2023smoothquant} and MASQuant~\cite{hu2026masquant} mitigate activation outliers through calibration-time transformations on frozen models. Quantization-aware training approaches, including OCC~\cite{wang2025optimizing} and Outlier Fallback~\cite{zhang2025accurate}, address outliers during training via clipping, compensation, or mixed-precision execution. However, these methods are not directly suited for end-to-end FP4 MLLM training: PTQ approaches rely on static calibration statistics, while existing QAT techniques are largely designed for text-only LLMs or higher-bit regimes and typically apply uniform heuristics across modules. In contrast, our findings show that the FP4 failure modes in MLLMs are both model-dependent and module-dependent, with activation outliers evolving dynamically during training and concentrating in different components in different architectures.

These observations suggest that stable FP4 training for MLLMs requires fully differentiable and training-aware activation-error modeling. To this end, motivated by~\cite{zhang2025accurate}, we propose \emph{Residual Fallback Quantization} (\emph{RFQ}), a lightweight activation-aware quantization strategy designed for ultra-low-bit MLLM training. RFQ decomposes activations into a quantized primary FP4 representation and an explicit quantization residual without any threshold-based selection. Instead of discarding or selectively preserving values, RFQ computes the full quantization error and reconstructs it through a secondary low-bit quantization pathway. This ensures that both the primary activations and residual correction remain compatible with FP4 computation. RFQ directly addresses FP4 sensitivity in MLLMs while requiring no architectural modifications or full-precision fallback.

\vspace{0.05in}
Our contributions are summarized as follows:
\begin{itemize}
\item We perform the first systematic study of ultra-low precision quantization in multimodal large language models, analyzing the behavior of MXFP8, MXFP4, and HiF4 across both multimodal generation and reasoning architectures, including Wan2.2 and Qwen3-VL.

\item We show that FP4 sensitivity is highly module-dependent in MLLMs, with visual components generally exhibiting greater susceptibility to quantization-induced degradation than language components. We further identify activation quantization as the dominant source of FP4 degradation, revealing that performance collapse is primarily driven by heavy-tailed activation distributions and activation outliers rather than weight quantization.

\item We propose \emph{Residual Fallback Quantization} (\emph{RFQ}), a lightweight activation-aware quantization strategy that compensates for FP4 quantization errors by re-quantizing activation residuals and accumulating their contributions during GEMM. RFQ effectively mitigates the accuracy degradation introduced by quantization while preserving FP4 computation for the majority of operations.

\item Extensive experiments demonstrate that RFQ substantially recovers the accuracy degradation introduced by MXFP4 and HiF4 quantization, achieving near BF16 baseline performance while preserving the efficiency benefits of ultra-low precision computation. 

\end{itemize}

\vspace{-0.2in}
\section{Related Work}
\vspace{-0.1in}
\noindent
\textbf{Quantization of LLMs.}
Quantization techniques are broadly split into post-training quantization (PTQ) and quantization-aware training (QAT). PTQ compresses a pretrained model using a small calibration dataset without further parameter optimization, making it a popular deployment choice due to low computational overhead. Representative methods include RTN~\cite{krishnamoorthi2018quantizing}, GPTQ~\cite{frantar2022gptq}, AWQ~\cite{lin2024awq}, and SmoothQuant~\cite{xiao2023smoothquant}, with recent extensions exploring outlier-aware allocation~\cite{lee2024owq}, mixed-precision~\cite{feng2025plmq}, and rotation-based transformations~\cite{he2025base}. However, PTQ struggles in ultra-low-bit regimes (e.g., 4-bit) because it cannot adapt parameters to quantization-induced distribution shifts.

In contrast, QAT integrates quantization into the training phase, allowing models to adapt to low-precision noise via gradient-based optimization. Recent frameworks like LLM-QAT~\cite{liu2024vptq}, BitDistiller~\cite{du2024bitdistiller}, EfficientQAT~\cite{chen2025efficientqat}, and GETA~\cite{qu2025automatic} leverage distillation and staged optimization to improve stability, while QuEST~\cite{panferov2025quest}, DB-LLM~\cite{chen2024db}, and BitNet~\cite{ma2025bitnet} introduce alternative ternary/binary parameterizations. Despite this, ultra-low precision QAT remains challenging due to activation outliers and non-uniform layer sensitivities, which introduce massive quantization noise and amplify error propagation across deep transformer blocks during backpropagation.

\noindent
\textbf{Quantization of Multimodal LLMs.}
Multimodal large language models (MLLMs) extend LLMs by integrating vision encoders, language backbones, and cross-modal fusion modules to enable joint reasoning over heterogeneous inputs such as images and text. However, this multimodal design introduces additional quantization challenges due to substantial differences in activation distributions across modalities and model components. As a result, direct application of LLM quantization techniques often leads to uneven degradation across modalities. To address this issue, recent PTQ methods introduce modality- and layer-aware designs. MBQ~\cite{li2025mbq} leverages modality-specific token sensitivities to improve calibration quality. LUQ~\cite{bhatnagar2025luq} studies ultra-low-bit PTQ for MLLMs and highlights strong layer-wise variation in quantization robustness driven by heterogeneous activation statistics. MQuant~\cite{yu2025mquant} further shows that cross-modal distribution mismatch is a key factor in quantization degradation. MASQuant~\cite{hu2026masquant} identifies smoothing imbalance across modalities, where dominant activation scales can suppress others under channel-wise smoothing, and proposes modality-aware compensation. Other approaches, including Q-VLM~\cite{wang2024q}, VLMQ~\cite{xue2025vlmq}, and Quant Experts~\cite{jia2026quant}, incorporate sensitivity-aware rounding, token-level importance, or reconstruction-based correction to reduce quantization error in vision-language models. Beyond PTQ, QAT for MLLMs remains relatively underexplored but increasingly important. Attn-QAT~\cite{zhang2026attn} analyzes FP4 attention training and identifies key stability constraints in low-precision backward computation. MF-QAT~\cite{xu2026mf} proposes multi-format QAT, enabling a single model to operate across multiple precision formats without retraining. 

Despite these advancements, stable optimization of MLLMs under ultra-low precision remains an open challenge due to dynamically evolving activation distributions and strong cross-modal interactions. Existing PTQ methods rely on static calibration that cannot adapt to shifting activations, while QAT approaches fail to model reconstruction under extreme bit constraints. These limitations are heavily amplified by the dynamic, modality-dependent nature of MLLMs statistics. To address this, we propose a unified, modality-agnostic quantization strategy that operates directly at the activation level, correcting quantization-induced distortions via differentiable reconstruction. By targeting errors at the activation level, our approach eliminates complex, modality-specific engineering while maintaining robustness under highly heterogeneous distributions.

\vspace{-0.2in}
\section{Preliminaries}
\vspace{-0.1in}
\subsection{Standard Floating-Point and Quantization Basics}

A standard floating-point ($FP$)~\cite{barrett1989formal} number is represented by a sign bit $s$, an exponent $e$, and a mantissa $m$. For a format with $E$ exponent bits and $M$ mantissa bits, the real-valued representation is given by:
\begin{equation}
v = (-1)^s \times 2^{e - \text{bias}} \times \left(1 + \frac{m}{2^M}\right),
\end{equation}
where $\text{bias}$ is the exponent bias and $\frac{m}{2^M}$ denotes the fractional contribution of the mantissa. Standard uniform quantization maps a real-valued tensor $x \in \mathbb{R}$ to a discrete set of low-precision values using a scaling factor $S$ and integer clipping bounds $[q_{\min}, q_{\max}]$:
\begin{equation}
\hat{x} = S \cdot \text{clip}\left(\text{round}\left(\frac{x}{S}\right), q_{\min}, q_{\max}\right),
\end{equation}
where $\text{round}(\cdot)$ denotes round-to-nearest integer rounding. While per-tensor and per-channel scaling strategies are effective for 8-bit integer quantization ($\mathrm{INT8}$), they become less reliable at ultra-low bit-widths (e.g., 4-bit). In such regimes, severe dynamic range mismatches and activation outliers lead to significant quantization error, making accurate representation difficult under fixed uniform scaling.

\subsection{Microscaling (MX) Block Specifications}
To mitigate accuracy degradation in ultra-low bit regimes, the OCP Microscaling Formats (MX)~\cite{rouhani2023microscaling} specification introduces a block-based scaling mechanism. Instead of applying a single scale factor to an entire tensor, elements are partitioned into small independent blocks of size $B$ (typically $B = 32$).
Within each block, all elements share a common scaling factor derived from a shared exponent. Given a block of high-precision values $\mathbf{X} = \{x_1, x_2, \dots, x_B\}$, the block scale is computed as:
\begin{equation}
S_{\text{block}} = 2^{\left\lfloor \log_2 \left(\max_i |x_i|\right) \right\rfloor}.
\end{equation}

Each element is then normalized by the block scale and quantized into a low-bit representation:
\begin{equation}
\tilde{x}_i = \mathcal{Q}_{\text{format}}\left(\frac{x_i}{S_{\text{block}}}\right),
\end{equation}
where $\mathcal{Q}_{\text{format}}(\cdot)$ denotes element-wise quantization to a low-bit floating-point or integer code.

\vspace{-0.2in}
\subsubsection{MXFP8 (Microscaling 8-bit Floating Point)}
The MXFP8 format defines two variants, E4M3 and E5M2, within the microscaling framework to balance precision and dynamic range. The E4M3 variant consists of 1 sign bit, 4 exponent bits, and 3 mantissa bits, providing higher precision at the cost of a narrower dynamic range, making it well suited for representing activations. In contrast, the E5M2 variant uses 1 sign bit, 5 exponent bits, and 2 mantissa bits, offering a wider dynamic range and improved robustness to large-magnitude variations.

\vspace{-0.2in}
\subsubsection{MXFP4 (Microscaling 4-bit Floating Point)}
MXFP4 further reduces the element-wise representation to 4 bits, typically using an E2M1 configuration (1 sign bit, 2 exponent bits, 1 mantissa bit). Due to the extremely limited precision, MXFP4 relies heavily on the shared block exponent $S_{\text{block}}$ to adaptively align the limited representable range with the local distribution of each tensor block.
\vspace{-0.1in}
\subsection{HiF4 (HiFloat4)}
While standard microscaling formats employ a flat, single-level block scaling factor, $\mathrm{HiF4}$~\cite{luo2026hifloat4,taghian2026hifloat4} introduces a multi-level hierarchical scaling paradigm designed for hardware-efficient acceleration. HiF4 organizes data into 64-element blocks with 32 bits of shared metadata, resulting in an amortized overhead of 0.5 bits per value (4.5 bits total per element).
The scaling hierarchy consists of a global base scale and two levels of binary micro-exponents. The global scale is represented using an unsigned 8-bit $\mathrm{E6M2}$ floating-point format with an exponent bias of 48, defining a coarse block-level normalization factor:
\begin{equation}
S_{\text{macro}} = 2^E \cdot (1.M).
\end{equation}

Fine-grained dynamic range refinement within the block is achieved using two tiers of 1-bit micro-exponents, $E1\_8$ (an 8-element vector) and $E1\_16$ (a 16-element vector). These micro-exponents provide localized exponent corrections over 8-element and 4-element sub-groups, respectively, effectively mitigating the impact of outliers and suppressing quantization noise.

Each individual element within the 64-element block is encoded using a 4-bit $\mathrm{S1P2}$ sign-magnitude format (1 integer bit and 2 fractional bits), conceptually equivalent to an $E1M2$ representation. The reconstructed value for the $i$-th element ($i \in [1, 64]$) is computed as:
\begin{equation}
V_i = S_{\text{macro}} \times 2^{E1\_8_{\lceil i/8 \rceil} + E1\_16_{\lceil i/4 \rceil}} \times \mathrm{S1P2}_i.
\end{equation}

This hierarchical formulation significantly expands the intra-block dynamic range to 4.81 binades while maintaining ultra-low precision storage. This capability is particularly critical for MLLMs; despite the use of Quantization-Aware Training, severe activation outliers still persistently emerge in these architectures. Consequently, the primary focus of our work is to leverage this hierarchical scaling to effectively mitigate these emergent outliers and preserve model accuracy.

\vspace{-0.1in}
\section{Our Approach}
\vspace{-0.1in}
\label{sec:method}
We study quantization-aware training (QAT) for multimodal LLMs under ultra-low-bit numerical formats. Our goal is to enable efficient deployment using aggressive quantization schemes such as MXFP4 and HiF4 while maintaining stable multimodal generation and reasoning performance. Therefore, we first construct a mixed-precision QAT framework tailored for MLLMs, where different model components are assigned heterogeneous numerical formats based on their quantization sensitivity. We then conduct a systematic diagnostic analysis to identify the dominant sources of degradation across modalities, layers, and tensor types (weights versus activations). Guided by these findings, we propose Residual Fallback Quantization (RFQ), an approach that mitigates activation-induced errors in ultra-low-bit regimes.

\vspace{-0.1in}
\subsection{Exploration of Ultra-Low-Bit Mixed-Precision QAT for MLLMs}
We investigate quantization-aware training of MLLMs using block-wise floating-point formats defined by the Open Compute Project (OCP) MX specification, including MXFP8 and MXFP4, together with the hierarchical HiF4 representation. To evaluate the effectiveness of these formats in large-scale multimodal settings, we conduct QAT experiments on two representative MLLMs: the generative video model \textit{Wan2.2 5B}~\cite{wan2025wanopenadvancedlargescale} and the reasoning-oriented model \textit{Qwen3-VL 30B}~\cite{Qwen3-VL}.

For our preliminary experiments, we adopt a mixed-precision QAT strategy that allocates numerical formats according to the quantization sensitivity of different model components. Specifically, the feed-forward networks (FFNs) and linear projection layers within both the vision and language modules are quantized using low-bit formats (MXFP8, W4A8, MXFP4, or HiF4), whereas components that are empirically more sensitive to precision reduction, such as embedding layers and the language modeling head, remain in BF16. This design aims to maximize compression efficiency while preserving multimodal generation and reasoning capabilities.

Our experiments reveal a clear separation in performance between 8-bit and ultra-low-bit 4-bit quantization regimes. As shown in Table~\ref{tab:QAT_training_loss}, MXFP8 preserves performance close to the BF16 baseline after QAT. We further observe that a mixed W4A8 configuration where weights are compressed to a 4-bit format while activations remain in 8-bit MXFP8 introduces only marginal additional degradation. In contrast, uniformly quantizing both weights and activations to 4-bit formats (MXFP4 or HiF4) results in substantially higher training loss. These findings suggest that MLLMs exhibit heterogeneous quantization sensitivity across tensors, with activations appearing considerably more vulnerable to aggressive precision reduction than weights.

Despite this careful allocation of precision, fully 4-bit quantization still leads to significant degradation, suggesting that format selection alone cannot bridge the performance gap. This points to the underlying activation distributions as a key source of quantization error, motivating a more fine-grained analysis of activation behavior in the following subsection.

\begin{algorithm}[t]
\caption{Residual Fallback Quantization (RFQ) GEMM}
\label{alg:rfq_gemm}
\scriptsize
\KwIn{
Input activation $X$, weight $Y$,
FP4 quantizer $\mathcal{Q}$,
fallback indicator $\phi(p,r)$,
block sizes $[P_s, Q_s, R_s]$
}

\KwOut{Output $Z$}

Partition $X$ into blocks $X^{p,r}$ and $Y$ into blocks $Y^{r,q}$\;

\For{$p = 0$ to $\lceil P/P_s \rceil - 1$}{
    \For{$q = 0$ to $\lceil Q/Q_s \rceil - 1$}{
        
        $Z^{p,q} \leftarrow 0$ \tcp*{Initialize block accumulator}
        
        \For{$r = 0$ to $\lceil R/R_s \rceil - 1$}{
            
            $\tilde{X}^{p,r} \leftarrow \mathcal{Q}(X^{p,r})$\;
            $\tilde{Y}^{r,q} \leftarrow \mathcal{Q}(Y^{r,q})$\;

            $Z^{p,q} \mathrel{+}= \tilde{X}^{p,r} \tilde{Y}^{r,q}$ \tcp*{Base FP4 GEMM}

            \If{$\phi(p,r) = 1$}{
                
                $\Delta X^{p,r} \leftarrow X^{p,r} - \tilde{X}^{p,r}$ \tcp*{Compute RFQ residual error}
                $\widehat{X}^{p,r} \leftarrow \mathcal{Q}(\Delta X^{p,r})$ \tcp*{Quantize residual to FP4}

                $Z^{p,q} \mathrel{+}= \widehat{X}^{p,r} \tilde{Y}^{r,q}$ \tcp*{Accumulate RFQ correction}
            }
        }
    }
}

\Return $Z$\;
\end{algorithm}

\begin{table}[h]
\centering
\scriptsize
\vspace{-0.2in}
\caption{Relative training loss increase (\%) for the BF16 baseline after QAT on models trained with over 5B tokens. MXFP8 and W4A8 show minimal degradation, while MXFP4 and HiF4 incur larger errors.}
\label{tab:QAT_training_loss}
\begin{tabular}{l|cccc}
\toprule
Model & MXFP8 & W4A8 & MXFP4 & HiF4 \\
\midrule
Wan2.2     & 0.30 & 0.80 & 7.10 & 2.70 \\
Qwen3-VL & 0.20 & 0.70 & 7.23 & 3.80 \\
\bottomrule
\end{tabular}
\vspace{-0.3in}
\end{table}

\vspace{-0.1in}
\subsection{Cross-Modal Weights–Activation Sensitivity}

\begin{wrapfigure}{r}[10pt]{0.60\columnwidth}
    \centering
    \includegraphics[width=\linewidth]{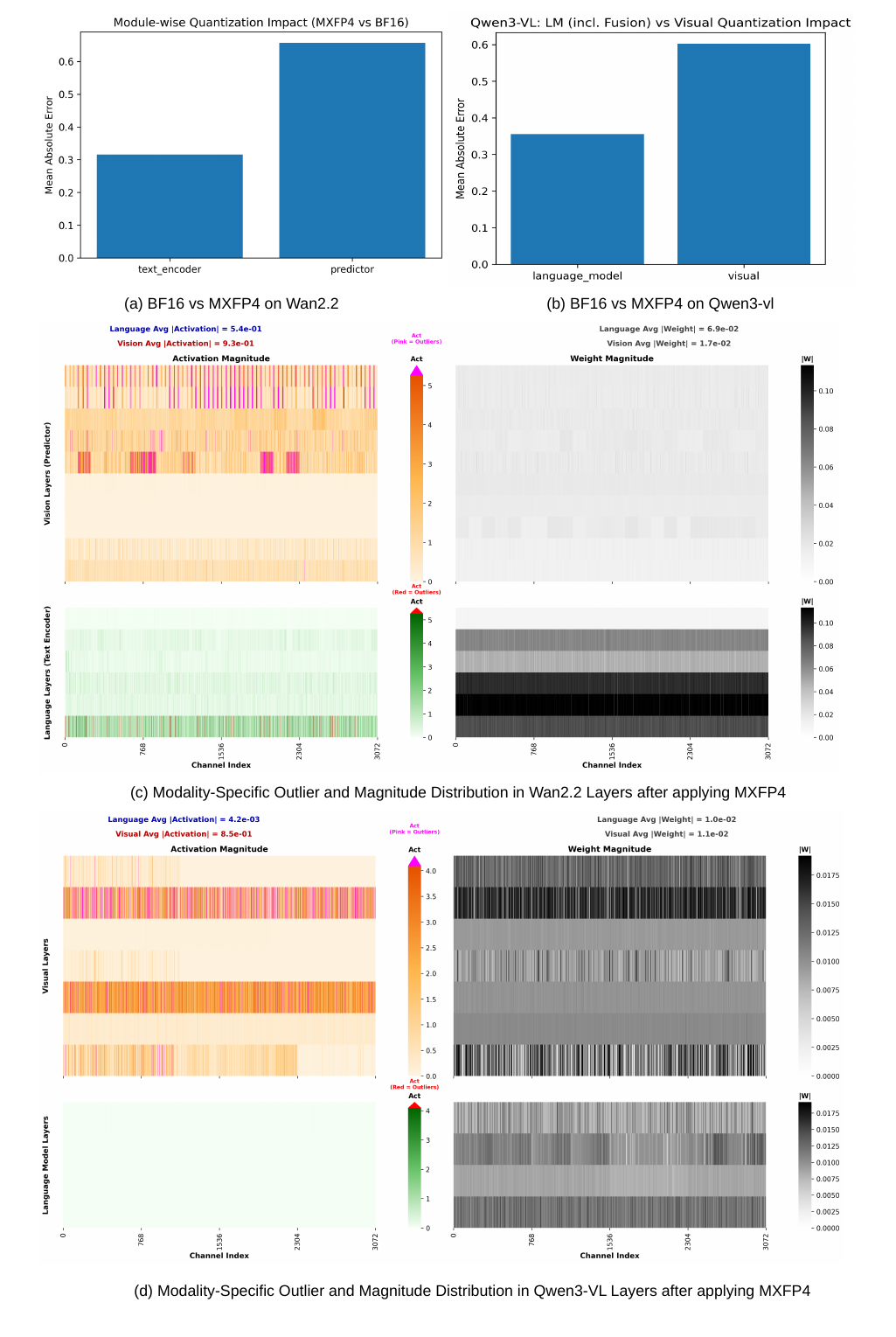}
    \vspace{-0.2in}
    \caption{Cross-modal sensitivity analysis under ultra-low-bit quantization.}
    \vspace{-0.15in}
    \label{fig:sensitivity}
\end{wrapfigure}

We further investigate whether the degradation under ultra-low-bit quantization is uniformly distributed across modalities or is primarily driven by a specific modality. In particular, we analyze the relative sensitivity of the vision and language components in both models under identical quantization configurations. To this end, we perform a controlled sensitivity study where vision and language pathways are quantized in the same numerical formats, allowing us to attribute performance variations to each modality independently. We conduct experiments on Wan2.2 and Qwen3-VL, both of which exhibit tightly coupled vision-language interactions during generation and reasoning. In prior work such as MBQ~\cite{Li_2025_CVPR}, where vision tokens are generally less sensitive than language tokens under post-training quantization, we observe a reverse trend in our QAT setting. Specifically, vision components are more sensitive to aggressive 4-bit quantization (MXFP4 and HiF4), resulting in greater degradation in both generative quality and reasoning consistency. This pattern is consistent on both models, although the effect is more pronounced in Wan2.2, which relies heavily on fine-grained visual detail reconstruction (see Figure~\ref{fig:sensitivity} (a) and (b)).
\setlength{\columnsep}{4pt}
\setlength{\intextsep}{3pt}

To better understand the source of this degradation, we analyze quantization sensitivity across weights and activations. As shown in Figure~\ref{fig:sensitivity}(c) and (d), activation tensors exhibit substantially larger dynamic ranges and more heterogeneous distributions compared to weights. This effect is particularly pronounced in vision-related layers, where extreme outliers significantly expand the quantization range under FP4 precision. As a result, the effective precision allocated to the majority of activation values is reduced, leading to distortion and frequent underflow of small-magnitude activations, which introduces substantial quantization residuals. In contrast, weight distributions remain comparatively compact and well-behaved, making them more amenable to low-bit quantization.

Overall, these findings indicate that the dominant failure mode in ultra-low-bit MLLM quantization is primarily driven by activation quantization errors shared in both language and vision pathways. Since both modalities are subject to severe precision constraints in activation propagation, weight-only mitigation strategies are insufficient to recover the resulting information loss. This observation motivates the need for activation-centric correction mechanisms. To this end, we propose Residual Fallback Quantization (RFQ). Inspired by the block-level fallback mechanism of~\cite{zhang2025accurate}, RFQ extends this idea to ultra-low-bit QAT by re-quantizing activation residuals through an efficient FP4-to-FP8 pathway, thereby improving accuracy while maintaining the computational efficiency of FP4 operations.

\vspace{-0.1in}
\subsection{Residual Fallback Quantization (RFQ)}

Motivated by our observation that activation quantization constitutes the dominant source of degradation in ultra-low-bit MLLMs, we propose \emph{Residual Fallback Quantization (RFQ)}, a residual-based correction mechanism that improves activation reconstruction while preserving the efficiency of uniform low-bit computation.

\noindent
Consider a matrix multiplication operation
\begin{equation}
Z = XY,
\end{equation}
where $X \in \mathbb{R}^{P \times R}$ denotes input activations and $Y \in \mathbb{R}^{R \times Q}$ represents the weights. Following a block-wise execution scheme, we partition $X$ into blocks $X^{p,r} \in \mathbb{R}^{P_s \times R_s}$ and $Y$ into blocks $Y^{r,q} \in \mathbb{R}^{R_s \times Q_s}$.

For each activation and weight block, we first perform conventional low-bit quantization using a target FP4 quantizer $\mathcal{Q}(\cdot)$:
\begin{equation}
\tilde{X}^{p,r} = \mathcal{Q}(X^{p,r}), \qquad
\tilde{Y}^{r,q} = \mathcal{Q}(Y^{r,q}).
\end{equation}

Although this base FP4 approximation is effective for the majority of tokens, severe activation outliers introduce substantial reconstruction errors. To selectively compensate for these errors without incurring global overhead, RFQ introduces a hardware-friendly fallback indicator $\phi(p,r) \in \{0, 1\}$, which identifies activation blocks containing significant outlier magnitudes. For blocks flagged by the indicator ($\phi(p,r) = 1$), RFQ computes the quantization residual:
\begin{equation}
\Delta X^{p,r} = X^{p,r} - \tilde{X}^{p,r}.
\end{equation}

Rather than storing this residual in a costly higher-precision format, we quantize it using the same underlying FP4 representation:
\begin{equation}
\widehat{X}^{p,r} = \mathcal{Q}(\Delta X^{p,r}).
\end{equation}

The final output block $Z^{p,q}$ is obtained by conditionally incorporating the residual correction term alongside the base GEMM computation:
\begin{equation}
Z^{p,q} = \sum_{r} \tilde{X}^{p,r}\tilde{Y}^{r,q} + \sum_{r} \phi(p,r) \cdot \left( \widehat{X}^{p,r}\tilde{Y}^{r,q} \right).
\label{eq:rfq}
\end{equation}

Equivalently, RFQ can be interpreted as dynamically approximating sensitive activation blocks using a two-stage low-bit decomposition:
\begin{equation}
X^{p,r} \approx \tilde{X}^{p,r} + \phi(p,r) \cdot \widehat{X}^{p,r},
\end{equation}
where both terms are restricted to the same low-bit numerical format. The first term captures the dominant signal component across all blocks, while the conditional residual term recovers precision lost to severe outliers only where strictly necessary.

\begin{figure*}[t]
    \centering
    \includegraphics[width=0.78\textwidth]{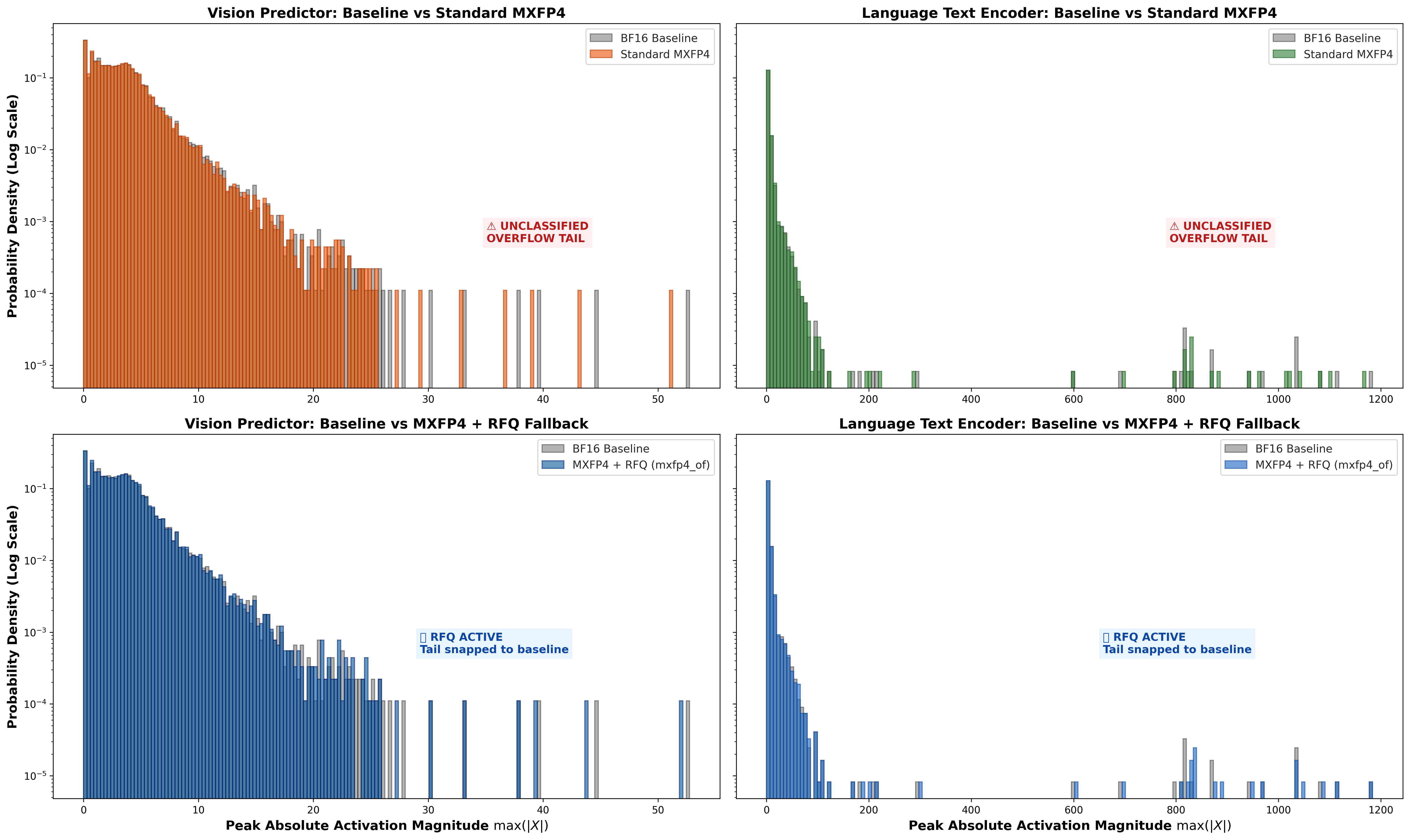}
    \caption{Comparative analysis of activation tail tracking and fallback interception across model modalities. Top Row highlights the unmitigated quantization noise and overflow elements in standard MXFP4 compared against the $BF16$ baseline. Bottom Row demonstrates how Residual Fallback Quantization (RFQ) restores fidelity by snapping extreme outliers.}
    \label{fig:rfq_histogram}
    \vspace{-0.25in}
\end{figure*}
Compared with global double-quantization alternatives, RFQ preserves the advantages of uniform low-bit arithmetic and triggers additional residual accumulation of GEMM when $\phi(p,r)=1$. Unlike traditional fallback frameworks that explicitly load pre-computed residual tensors from off-chip DRAM inside the execution loop (e.g., $u(i,k)$ in baseline FQ~\cite{dettmers2022llmint8, zhang2025accurate}), RFQ operates as a streaming kernel that evaluates both the residual error $\Delta X^{p,r}$ and its subsequent quantization of FP4 $\widehat{X}^{p,r}$ completely on-the-fly. Furthermore, by restricting both the base and residual pathways to the exact same uniform FP4 format $\mathcal{Q}(\cdot)$ rather than heterogeneous integer types, RFQ minimizes hardware execution paths. Crucially, this residual correction mechanism is applied exclusively during the forward pass to avoid compounding memory and gradient overhead during backward propagation, keeping the implementation fully compatible with existing low-bit hardware tensor primitives and standard quantization-aware training infrastructure. The complete execution flow is summarized in Algorithm~\ref{alg:rfq_gemm} and the RFQ effect is illustrated in Figure~\ref{fig:rfq_histogram}.

\begin{table}[t]
\centering
\caption{Parameter breakdown (\%) by functional component with respect to the total paramter size. Quant.~(\%) for the visual and textual modules denotes the proportion of parameters quantized within each module. Projection layers and mergers are included in the ``Other'' category.}
\label{tab:merged_parameter_breakdown}
\scriptsize
\setlength{\tabcolsep}{3pt}
\renewcommand{\arraystretch}{1.05}
\begin{tabular}{llcccc|c}
\toprule
\multirow{2}{*}{\textbf{Model}} &
\multirow{2}{*}{\textbf{Module}} &
\multicolumn{4}{c|}{\textbf{Parameter Distribution (\%)}} &
\multirow{2}{*}{\textbf{Quant.(\%)}} \\
\cmidrule(lr){3-6}
& & \textbf{Atten. Proj.} & \textbf{FFN/MLP} & \textbf{MoE} & \textbf{Other} & \\
\midrule
\multirow{3}{*}{Wan2.2}
& Visual  & 39.86 & 46.51 & -- & 11.62  & 97.99 \\
& Textual & 28.35 & 53.16 & -- & -- & 81.50 \\
\midrule
\multirow{3}{*}{Qwen3-VL}
& Visual  & 26.61 & 49.71 & -- & 22.77  & 99.09 \\
& Textual & 2.96 & -- & 94.95 & --  & 97.91 \\

\bottomrule
\end{tabular}
\vspace{-0.1in}
\end{table}


\vspace{-0.1in}
\section{Experimental Analysis}
\vspace{-0.05in}
\begin{table*}[t]
\centering
\scriptsize
\setlength{\tabcolsep}{4pt}
\renewcommand{\arraystretch}{1.1}
\caption{For Wan2.2, quantitative comparison on VBench. Values in parentheses indicate the relative change (\%) with respect to the BF16 baseline. Blue denotes improvement and red denotes degradation.}
\label{tab:video_gen}

\newcommand{\vertheader}[1]{\rotatebox{90}{\parbox{0.55in}{\raggedright #1}}}
\begin{tabular}{lcccccccc}
\toprule
Method &
\vertheader{Subject Consistency} &
\vertheader{Background Consistency} &
\vertheader{Imaging Quality} &
\vertheader{Temporal Flickering} &
\vertheader{Motion Smoothness} &
\vertheader{Dynamic Degree} &
\vertheader{Overall Consistency} &
\vertheader{Aesthetic Quality} \\
\midrule

Baseline (BF16) &
95.74 & 96.77 & 65.32 & 98.42 & 99.25 & 45.00 & 7.03 & 59.43 \\
\midrule

MXFP4 &
95.65 & 96.60 & 66.69 & 98.10 & 99.15 & 43.00 & \textbf{7.24} & 58.93 \\
&
{\tiny \down{0.09}} &
{\tiny \down{0.17}} &
{\tiny \up{1.37}} &
{\tiny \down{0.32}} &
{\tiny \down{0.10}} &
{\tiny \down{2.00}} &
{\tiny \up{0.21}} &
{\tiny \down{0.5}} \\
\midrule

MXFP4 + RFQ (ours) &
95.43 & 96.54 & 66.48 & 98.29 & 99.23 & 50.00 & 6.90 & 59.26 \\
&
{\tiny \down{0.31}} &
{\tiny \down{0.23}} &
{\tiny \up{1.16}} &
{\tiny \down{0.13}} &
{\tiny \down{0.02}} &
{\tiny \up{5.00}} &
{\tiny \down{0.13}} &
{\tiny \down{0.16}} \\
\midrule

HiF4 &
95.23 & \textbf{96.67} & \textbf{67.36} & 98.33 & 99.24 & \textbf{53.00} & 6.98 & 59.44 \\
&
{\tiny \down{0.51}} &
{\tiny \down{0.10}} &
{\tiny \up{2.04}} &
{\tiny \down{0.09}} &
{\tiny \down{0.01}} &
{\tiny \up{8.00}} &
{\tiny \down{0.05}} &
{\tiny \up{0.01}} \\
\midrule

HiF4 + RFQ (ours) &
\textbf{95.86} & 96.60 & 66.79 & \textbf{98.38} & \textbf{99.24} & 51.00 & 6.86 & \textbf{59.54} \\
&
{\tiny \up{0.12}} &
{\tiny \down{0.17}} &
{\tiny \up{1.47}} &
{\tiny \down{0.04}} &
{\tiny \down{0.01}} &
{\tiny \up{6.00}} &
{\tiny \down{0.17}} &
{\tiny \up{0.19}} \\

\bottomrule
\end{tabular}
\vspace{-0.2in}
\end{table*}

\begin{table}[htbp]
\centering
\scriptsize
\caption{Accuracy comparison across different datasets and quantization formats on Qwen3-VL. Values in parentheses indicate the absolute change with respect to the BF16 baseline. Blue denotes improvement, while red indicates degradation.}
\vspace{-0.05in}
\label{tab:reasoning}
\begin{tabular}{lcccc}
\toprule
Method & RealWorldQA & MMStar & MMBenchEN & SimpleVQA \\
\midrule
BF16
& 72.68
& 70.80
& 90.77
& 16.83 \\
\midrule
MXFP4
& 70.98
& 69.67
& 90.72
& 15.16 \\
& {\tiny \down{1.70}}
& {\tiny \down{1.13}}
& {\tiny \down{0.05}}
& {\tiny \down{1.67}} \\
\midrule
MXFP4 + RFQ (ours)
& 72.16
& 70.73
& 90.40
& \textbf{16.44} \\
& {\tiny \down{0.52}}
& {\tiny \down{0.07}}
& {\tiny \down{0.37}}
& {\tiny \down{0.39}} \\
\midrule
HiF4
& 72.42
& 71.27
& 90.50
& 15.35 \\
& {\tiny \down{0.26}}
& {\tiny \up{0.47}}
& {\tiny \down{0.27}}
& {\tiny \down{1.48}} \\
\midrule
HiF4 + RFQ (ours)
& \textbf{72.81}
& \textbf{71.47}
& \textbf{90.77}
& 15.66 \\
& {\tiny \up{0.13}}
& {\tiny \up{0.67}}
& {\tiny (0.00)}
& {\tiny \down{1.17}} \\
\bottomrule
\vspace{-0.35in}
\end{tabular}
\end{table}

\subsection{Experimental Setup}
We evaluate our proposed RFQ framework under ultra-low-bit precision constraints on two representative multimodal architectures: Wan2.2-5B and Qwen3VL-30B. For both models, we initialize from publicly available pretrained checkpoints and subsequently perform low-precision quantization-aware fine-tuning.
For the Qwen3-VL supervised fine-tuning (SFT) phase, we utilize the concept-balanced CC-3M dataset comprising 595K samples~\cite{liu2023improvedllava, liu2023llava}. For Wan2.2, we leverage a subset of the OpenVid-1M dataset~\cite{nan2024openvid}, specifically utilizing Part 1 that contains approximately 26,000 video-text pairs. Following the mixed-precision scheme outlined in Section~\ref{sec:method}, the embedding layers and the final language modeling head are maintained in BF16 precision, whereas all remaining linear, MoE and attention projection layers are quantized using MXFP8, W4A8, MXFP4, or HiF4 formats. See Table.~\ref{tab:merged_parameter_breakdown} for the quantized parameter count. To ensure a fair and rigorous comparison across these diverse numerical configurations, both models are fine-tuned under identical quantization hyperparameters for approximately 5 billion tokens. See supplemental for more details. 

For both Wan2.2 and Qwen3-VL, we performed SFT using AdamW starting from their respective publicly available pretrained checkpoints. Wan2.2 is fine-tuned with a learning rate of $1\times10^{-5}$, while Qwen3-VL uses a learning rate of $1\times10^{-7}$. Unless otherwise specified, all other training hyperparameters remain unchanged in different quantization configurations to ensure fair comparisons. We evaluated Wan2.2 using VBench on 100 randomly sampled prompts from the MovieGen~\cite{polyak2024movie} benchmark and report downstream performance on different dimensions.
For Qwen3-VL, we assess multimodal reasoning capabilities on four widely used benchmarks: RealWorldQA~\cite{realworldqa2024}, MMStar~\cite{chen2024we}, MMBench-EN~\cite{liu2024mmbench}, and SimpleVQA~\cite{cheng2025simplevqamultimodalfactualityevaluation}. We report the VQA accuracy on each benchmark. All evaluations are conducted using the same inference protocol for both BF16 and quantized models. 

\vspace{-0.15in}
\subsection{Evaluation Performance}
\setlength{\columnsep}{10pt}

\begin{wrapfigure}{r}{0.50\textwidth}
\vspace{-0.3in}
    \centering
    \includegraphics[width=0.45\textwidth]{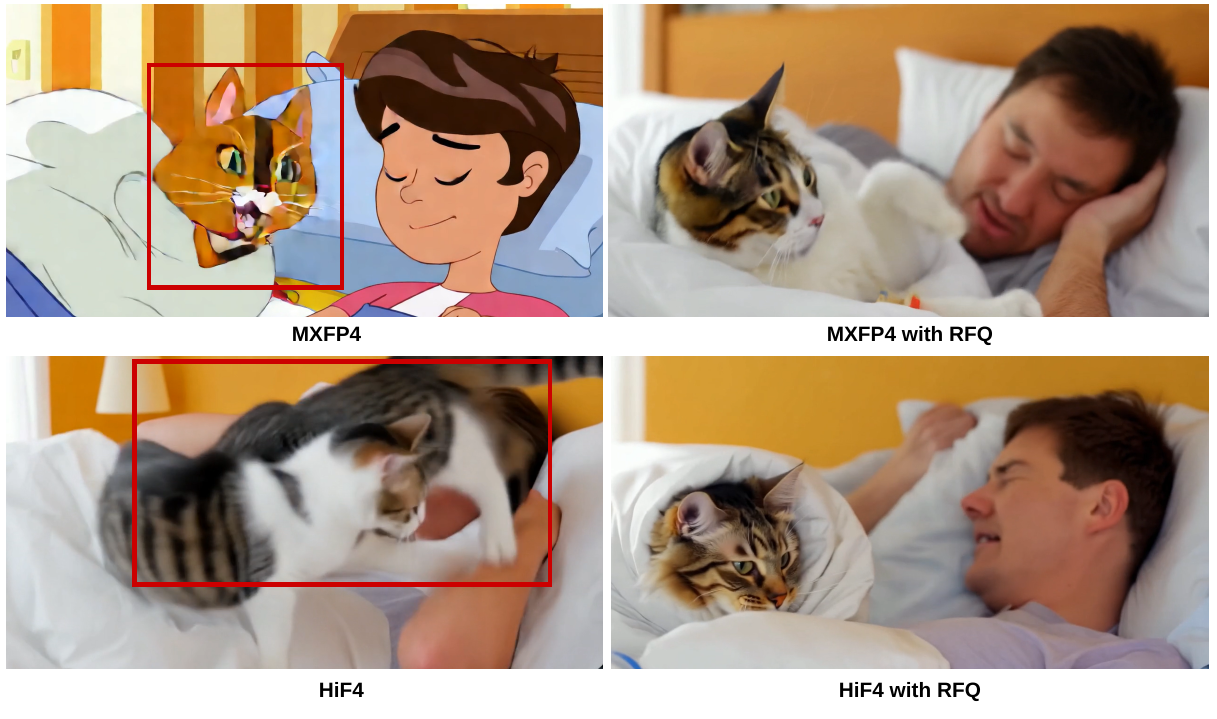}
    \caption{Qualitative results for video generation. While 4-bit quantization methods (\emph{i.e.}, MXFP4 and HiF4) degrade generation quality, our proposed RFQ recovers visual fidelity and improves temporal consistency.}
    \label{fig:qualatative_results}
\end{wrapfigure}

\textbf{Performance of video generation.} Table~\ref{tab:video_gen} presents quantitative results on Wan2.2 evaluated using VBench. Compared with standard MXFP4 quantization, RFQ consistently improves multiple video generation metrics. In particular, RFQ improves the Dynamic Degree from 43.00 to 50.00 while simultaneously improving Aesthetic Quality and maintaining competitive Subject and Background Consistency. Similar trends are observed in the HiF4 setting, where RFQ improves Subject Consistency from 95.23 to 95.86 and Aesthetic Quality from 59.44 to 59.54. We also include qualitative results in Figure~\ref{fig:qualatative_results}. See \textbf{supplemental} for more results. These results demonstrate that RFQ effectively mitigates the adverse effects of activation quantization in ultra-low-bit video generation models.

\noindent
\textbf{Performance of multimodal reasoning.}
Table~\ref{tab:reasoning} reports the results of Qwen3-VL. In MXFP4 quantization, RFQ consistently narrows the performance gap with respect to the baseline BF16, improving RealWorldQA from 70.98 to 72.16, MMStar from 69.67 to 70.73, and SimpleVQA from 15.16 to 16.44. Under the HiF4 configuration, RFQ achieves further gains, improving RealWorldQA from 72.42 to 72.81 (even better than BF16 baseline) and MMStar from 71.27 to 71.47, while recovering the BF16 performance in MMBench-EN. These findings indicate that the RFQ generalizes across various multimodal reasoning tasks and effectively improves the robustness of ultra-low-bit MLLMs.

\begin{figure}[t]
    \centering
    \includegraphics[width=0.95\textwidth]{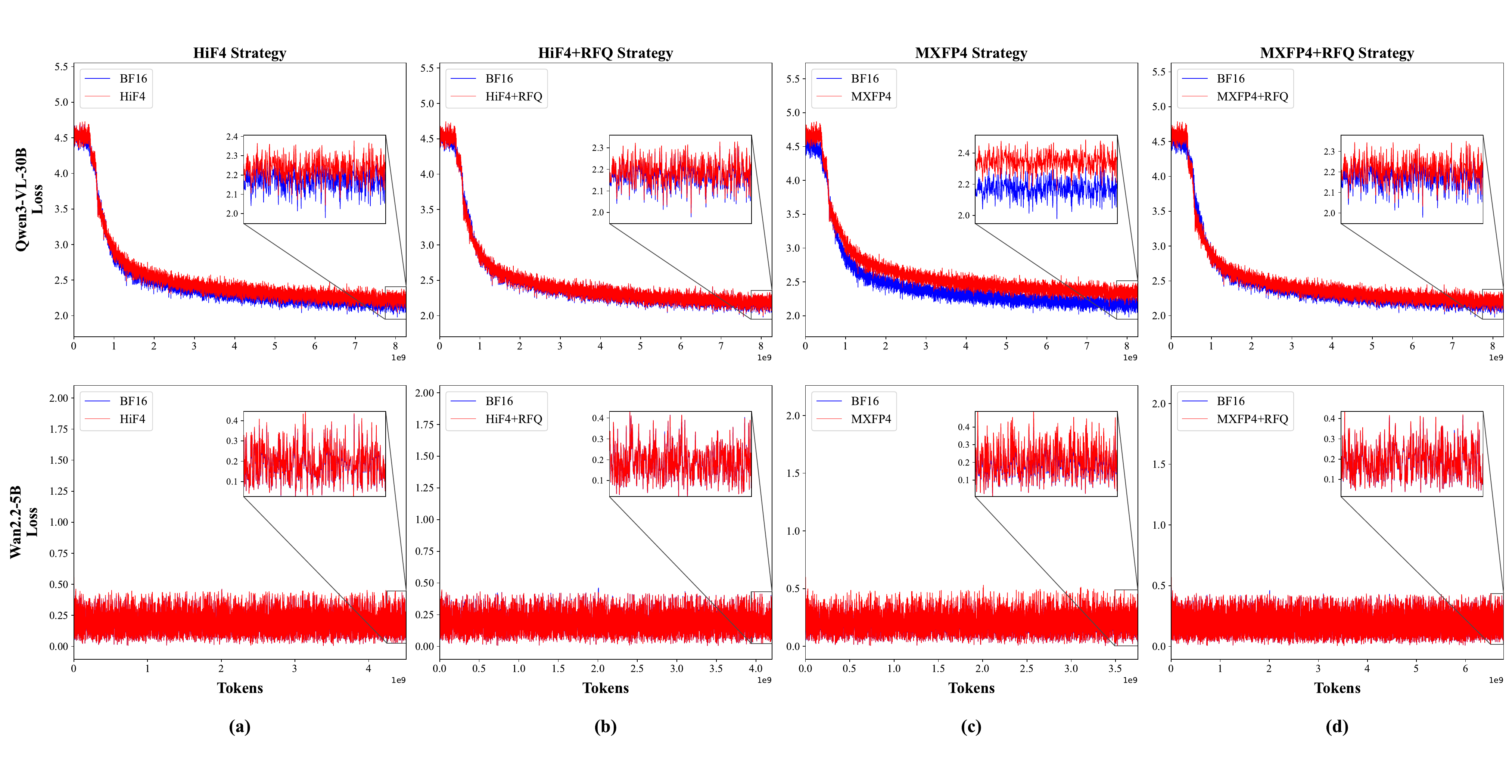}
      \vspace{-0.3in}
    \caption{Training loss curves comparing low-bit FP4 strategies against the BF16 baseline across Qwen3-VL-30B and Wan2.2-5B.}
    \label{fig:teaser}
    \vspace{-0.1in}
\end{figure}

In addition to downstream performance, Table~\ref{tab:QAT} reports the relative training loss difference with respect to the BF16 baseline. RFQ consistently enhances both generative and reasoning performance under aggressive FP4 quantization schemes, leading to improvements in both SFT and downstream evaluation. Notably, these gains are obtained without increasing the precision of the underlying representation, demonstrating the effectiveness of residual fallback correction in alleviating activation-induced quantization errors.

\begin{table}[t]
\centering
\scriptsize
\vspace{-0.05in}
\caption{Relative training loss increase (\%) over the BF16 baseline after QAT with 4-bit quantization. RFQ substantially mitigates the degradation caused by MXFP4 and HiF4.}
\vspace{-0.1in}
\label{tab:QAT}
\begin{tabular}{l|c|c|c|c}
\toprule
Model & MXFP4 & MXFP4 + RFQ (ours) & HiF4 & HiF4 + RFQ (ours) \\
\midrule
Wan2.2     & 7.50 & \textbf{1.14} & 2.89 & \textbf{0.61} \\
Qwen3VL-30B & 7.23 & \textbf{1.43} & 2.32 & \textbf{0.66} \\
\bottomrule
\end{tabular}
\vspace{-0.25in}
\end{table}
\vspace{-0.2in}
\section{Conclusion}
\vspace{-0.1in}
In this paper, we present a systematic study of ultra-low-bit supervised fine-tuning (SFT) for multimodal LLMs. Our analysis shows that visual modules are generally more sensitive to aggressive quantization than language modules, though quantizing language modules also leads to non-negligible performance degradation. These findings suggest that effective accuracy recovery in ultra-low-bit settings requires jointly addressing both components. Based on this observation, we propose RFQ, a simple yet effective approach that exploits residual quantization errors through an auxiliary uniform quantization pathway to compensate for information loss under ultra-low-bit quantization. Extensive experiments demonstrate that RFQ consistently improves training dynamics and downstream performance, recovering accuracy close to the BF16 baseline. Overall, our results highlight the promise of residual-based compensation for efficient deployment of multimodal foundation models under tight memory and computation constraints. In future work, we will explore more aggressive quantization-aware training settings, including 2-bit quantization, and extend RFQ to broader multimodal architectures and tasks.


{\small
\bibliographystyle{splncs04}
\bibliography{main}
}

\end{document}


\title{Activation Outliers Matter: Robust Recovery for Quantized Multimodal LLMs \linebreak--Supplemental--} 

\titlerunning{Robust Recovery for Quantized Multimodal LLMs - supp}


\author{Tanzila Rahman \and
Mehran Taghian Jazi \and
Yunke Peng \and
Zhuang Ma \and
Anandharaju Durai Raju \and
Yao Wang \and
Xing Huang \and
Hei Yi Mak \and
Shadan Golestan \and
Hoang Le \and
Yonghan Dong \and
Wei Guo \and
Yaoyuan Wang}

\authorrunning{T. Rahman et al.}


\institute{Huawei\\
\email{\{tanzila.rahman, pengyunke\}@huawei.com}}

\maketitle

\begin{figure}[!htbp]
    \centering
    \includegraphics[width=0.80\textwidth]{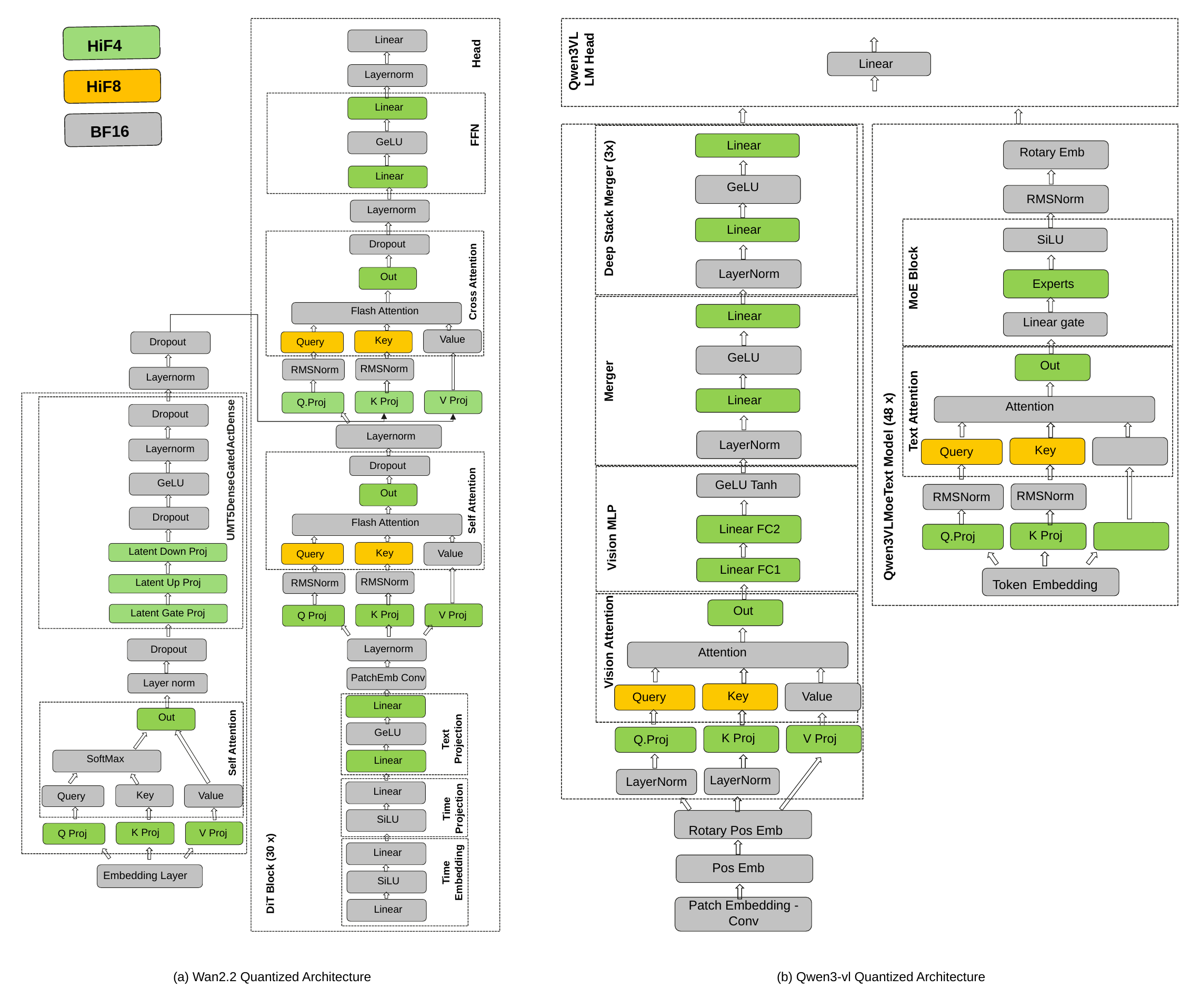}
    \caption{Overview of the proposed mixed-precision quantization framework for Wan2.2 and Qwen3-VL.}
    \label{fig:workflow}
\end{figure}

\noindent
\textbf{Mixed-Precision Architectural Mapping.} We apply ultra-low-bit quantization to the main computational components of the Wan2.2 and Qwen3-VL models. Specifically, we quantize the attention Query-Key-Value (QKV) projections, output projections, and the feed-forward experts in the Mixture-of-Experts (MoE) layers. Figure~\ref{fig:workflow} illustrates the mixed-precision design, highlighting which layers are quantized to 4-bit weights and which are preserved in BF16. To maintain model stability and accuracy, numerically sensitive components, including the embedding layers, output (LM) head, and other critical modules, remain in BF16 precision. In addition, we quantize the Query and Key activations to 8 bits before the FlashAttention operation, reducing memory bandwidth and improving inference efficiency. 

\noindent
\textbf{Additional Experimental Results.} Table~\ref{tab:video_gen1} presents the VBench evaluation results for Wan2.2, comparing the baseline model with MXFP8 and W4A8 quantization schemes. Notably, MXFP8 achieves performance comparable to or better than baseline in most evaluation dimensions, including subject consistency, background consistency, imaging quality, and motion smoothness. Similarly, W4A8 maintains competitive performance relative to the baseline, demonstrating that aggressive low-bit quantization can be applied to video generation models with minimal impact on output.
Table~\ref{tab:reasoning1} reports the reasoning performance of Qwen3-VL on four benchmark datasets. Both MXFP8 and W4A8 achieve results that closely match those of the baseline model, indicating that low-bit quantization preserves the reasoning capabilities of large multimodal language models. 

We further investigate a mixed-precision configuration in which the attention query-key (\textit{QK}) computation is quantized to 8-bit precision using \emph{HiF8} or \emph{MXFP8} before being processed by FlashAttention, while all linear layers, attention projections, feed-forward networks, and MoE experts remain quantized to 4-bit precision. The corresponding results are presented in Table~\ref{tab:video_gen2}. Despite the additional quantization applied to the attention mechanism, this configuration continues to achieve performance that is either comparable to or, in some cases, surpasses that of the baseline model, highlighting the robustness of the proposed mixed-precision quantization strategy.

\begin{table*}[t]
\centering
\scriptsize
\setlength{\tabcolsep}{4pt}
\renewcommand{\arraystretch}{1.1}
\caption{Quantitative Evaluation of MXFP8 and W4A8 Quantization on Wan2.2 Using VBench.}
\label{tab:video_gen1}
\newcommand{\vertheader}[1]{\rotatebox{90}{\parbox{0.55in}{\raggedright #1}}}
\begin{tabular}{lcccccccc}
\toprule
Method &
\vertheader{Subject Consistency} &
\vertheader{Background Consistency} &
\vertheader{Imaging Quality} &
\vertheader{Temporal Flickering} &
\vertheader{Motion Smoothness} &
\vertheader{Dynamic Degree} &
\vertheader{Overall Consistency} &
\vertheader{Aesthetic Quality} \\
\midrule

Baseline (BF16) &
95.74 & 96.77 & 65.32 & 98.42 & 99.25 & 45.00 & 7.03 & 59.43 \\
\midrule
MXFP8 &
95.82 & 96.88 & 66.38 & 98.44 & 99.28 & 50.00 & 6.96 & 59.02 \\
\midrule
W4A8 &
95.43 & 96.54 & 66.48 & 98.29 & 99.23 & 50.00 & 6.90 & 59.51 \\
\bottomrule
\end{tabular}
\end{table*}

\begin{figure}[!htbp]
    \centering
    \includegraphics[width=0.97\textwidth]{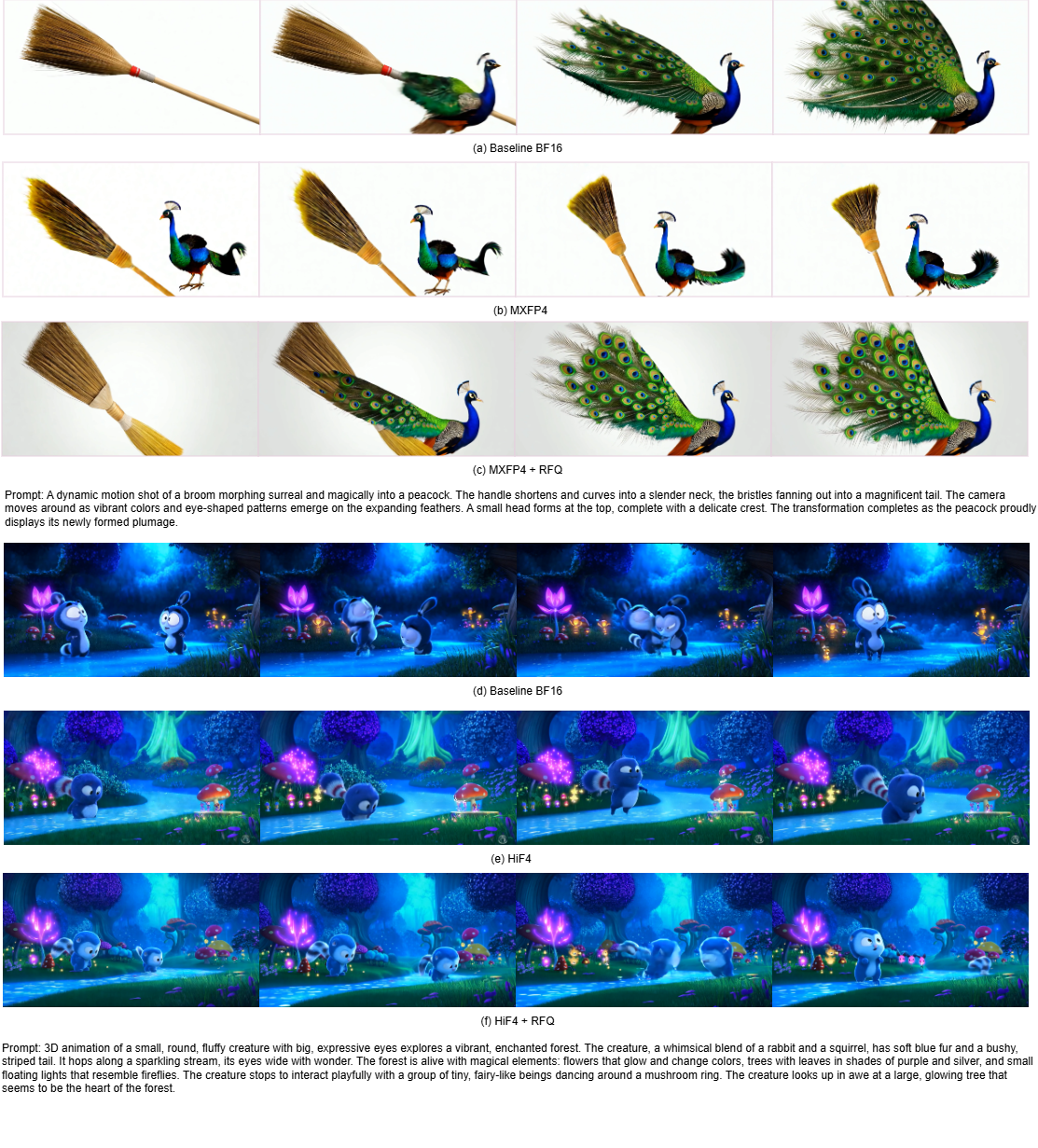}
    \caption{Qualitative comparison of Wan2.2 using BF16 and 4-bit quantization (MXFP4 and HiF4). RFQ effectively mitigates the quality degradation introduced by 4-bit quantization, producing results closer to the BF16 baseline.}
    \label{fig:qualittaive_wan}
\end{figure}

\begin{table}[t]
\centering
\scriptsize
\caption{Accuracy Comparison of MXFP8 and W4A8 Quantization on Qwen3-VL Across Multiple Benchmarks.}
\label{tab:reasoning1}
\begin{tabular}{lcccc}
\toprule
Method & RealWorldQA & MMStar & MMBenchEN & SimpleVQA \\
\midrule
BF16 & 72.68 & 70.80 & 90.77 & 16.83 \\
\midrule
MXFP8 & 73.33 & 69.93 & 90.86 & 16.49 \\
\midrule
W4A8 & 73.59 & 71.33 & 90.56 & 15.92 \\
\bottomrule
\end{tabular}
\end{table}

\begin{table*}[htbp]
\centering
\scriptsize
\setlength{\tabcolsep}{4pt}
\renewcommand{\arraystretch}{1.1}
\caption{Quantitative Evaluation of 4-bit quantization with 8-bit attention QK Quantization on Wan2.2 Using VBench.}
\label{tab:video_gen2}
\newcommand{\vertheader}[1]{\rotatebox{90}{\parbox{0.55in}{\raggedright #1}}}
\begin{tabular}{lcccccccc}
\toprule
Method &
\vertheader{Subject Consistency} &
\vertheader{Background Consistency} &
\vertheader{Imaging Quality} &
\vertheader{Temporal Flickering} &
\vertheader{Motion Smoothness} &
\vertheader{Dynamic Degree} &
\vertheader{Overall Consistency} &
\vertheader{Aesthetic Quality} \\
\midrule

Baseline (BF16) &
95.74 & 96.77 & 65.32 & 98.42 & 99.25 & 45.00 & 7.03 & 59.43 \\
\midrule
HiF4 + RFQ + Attn. QK HiF8 &
 95.50 & 96.73 & 65.56 & 98.31 & 99.24 & 48 & 7.07 & 59.44\\
 \midrule
 HiF4 + RFQ + Attn. QK MXFP8 &
 95.48 & 96.78 & 66.33 & 98.31 & 99.25 & 50 & 6.96 & 59.42 \\
\bottomrule
\end{tabular}
\end{table*}

Figure~\ref{fig:qualittaive_wan} shows an additional qualitative comparison of the results of the Wan2.2 video generation using the baseline BF16 model and 4-bit quantization schemes (MXFP4 and HiF4), evaluated both with and without the proposed Residual Fallback Quantization (RFQ). Direct 4-bit quantization introduces noticeable degradation in visual quality and temporal consistency. By contrast, incorporating RFQ substantially mitigates these quantization-induced artifacts, producing videos with improved visual fidelity, motion coherence, and overall generation quality, closely matching the BF16 baseline.

